\documentclass[preprint,12pt]{elsarticle}

\usepackage{amssymb}
\usepackage{amsmath}
\usepackage{tikz}
\usepackage{multirow}
\usepackage{adjustbox}
\usepackage{pstricks}
\usepackage{xcolor}

\journal{Nuclear Physics B}

\begin{document}

\begin{frontmatter}



\title{TenAd: A Tensor-based Low-rank Black Box Adversarial Attack for Video Classification}


\author[add1]{Kimia Haghjooei }
\ead{kimia.haghjooei@modares.ac.ir}
\author[add1]{Mansoor Rezghi\corref{mycorrespondingauthor}}
\ead{rezghi@modares.ac.ir}
\address[add1]{ Department of Computer Science, Tarbiat Modares University, Tehran, Iran.}
\cortext[mycorrespondingauthor]{Corresponding author}

\begin{abstract} Deep learning models have achieved remarkable success in computer vision but remain vulnerable to adversarial attacks, particularly in black-box settings where model details are unknown. Existing adversarial attack methods(even those works with key frames) often treat video data as simple vectors, ignoring their inherent multi-dimensional structure, and require a large number of queries, making them inefficient and detectable. In this paper, we propose \textbf{TenAd}, a novel tensor-based low-rank adversarial attack that leverages the multi-dimensional properties of video data by representing videos as fourth-order tensors. By exploiting  low-rank attack, our method significantly reduces the search space and the number of queries needed to generate adversarial examples in black-box settings. Experimental results on standard video classification datasets demonstrate that \textbf{TenAd} effectively generates imperceptible adversarial perturbations while achieving higher attack success rates and query efficiency compared to state-of-the-art methods. Our approach outperforms existing black-box adversarial attacks in terms of success rate, query efficiency, and perturbation imperceptibility, highlighting the potential of tensor-based methods for adversarial attacks on video models. \end{abstract}




\end{frontmatter}


\section{Introduction}
	
	Deep learning models have demonstrated significant success across a wide range of tasks, such as face recognition \cite{Taigman2014}; action recognition \cite{Carreira2017}; object detection— \cite{Ren2023}; and video surveillance for enhancing automated security analysis \cite{Redmon2023}. Despite these successes, recent studies have highlighted a critical vulnerability that these models are highly susceptible to adversarial examples \cite{ Malhotra2023}.

	Adversarial examples are carefully crafted inputs that involve minimal perturbations imperceptible to the human eye, yet they can effectively mislead deep neural networks (DNNs) into making incorrect predictions \cite{Szegedy2014,Goodfellow2015}. The susceptibility of DNN models to adversarial attacks raises serious concerns regarding their reliability and robustness, particularly when these models are employed in security-critical applications such as autonomous driving, biometric authentication, and medical diagnostics \cite{Dhamija2024, Almutairi2023, Vineetha2023}.
	
	Adversarial attacks can generally be divided into two main categories. The first is white-box attacks, where the attacker has full access to the target model’s internal information, including its parameters and objective function \cite{Akhtar2021, Chen2017, Brendel2017}. The second category is black-box attacks, where the attacker has no direct knowledge of the model’s architecture or parameters and can only interact with it by querying its predictions for given inputs \cite{Ran2025}. Black-box attacks are particularly relevant for real-world scenarios, where access to model details is typically restricted.
	
	Adversarial examples can also be generated in two distinct ways: targeted and untargeted. In a targeted attack, the goal is to mislead the model into making a specific incorrect prediction \cite{MoosaviDezfooli2016}. In contrast, untargeted attacks aim to force the model into producing any incorrect output, as long as it differs from the original prediction.
	
	Recent research on adversarial examples has primarily focused on image recognition models, employing both white-box and black-box strategies. However, the study of adversarial attacks on video recognition models, particularly in a black-box setting, remains relatively unexplored \cite{Wei2019}. Unlike images, videos are inherently high-dimensional, incorporating both spatial and temporal data, which greatly increases the complexity and search space for adversarial attacks \cite{Wei2019}. Given the widespread use of real-time video classification models, especially in critical applications such as surveillance, it is crucial to thoroughly investigate how these models respond to adversarial threats. Understanding these effects will help develop more robust defenses to protect against vulnerabilities in real-world settings \cite{Eleftheriadis2024}.
	
	Consider video data represented as $\mathcal{X} \in \mathbb{R}^{W \times H \times C \times T}$, where $T$ represents the number of frames, $C$ denotes the number of channels, and $W \times H$ defines the spatial dimensions of each frame. Finding the optimal perturbation for such high-dimensional data requires exploring a vast search space of $\mathbb{R}^{T \times C \times W \times H}$, which is computationally expensive and requires a significant number of queries, particularly in a black-box attack scenario.
	
	Given the inherent spatial and temporal redundancies in video data, a promising strategy to enhance the efficiency of adversarial attacks is to reduce the dimensionality of the problem by selecting a subset of frames, known as keyframes. These keyframes are crucial for video recognition tasks and play a pivotal role in video representation. Various techniques can be employed to select keyframes, such as reinforcement learning, as utilized in the SVA attack \cite{sva}; heuristic algorithms, as in the Heuristic attack \cite{heu}; or unsupervised methods key frame selection described in \cite{rezghi2023}. By targeting only keyframes, the complexity of the attack can be reduced substantially, making it more feasible to execute.
	
	Moreover, different regions within a video frame contribute unequally to the classification outcome. To further enhance the attack's effectiveness, another technique is to target only the most salient regions within a frame, which have the highest impact on classification results \cite{heu}. By focusing perturbations on these important regions, attackers can increase the efficiency of the attack while minimizing the number of changes needed.
	
	Despite these achievements, attackers still face a significant challenge in identifying the optimal set of keyframes and salient regions to perturb effectively. Even after reducing the complexity of the search space, black-box attacks still often require numerous queries, which is both time-consuming and can lead to increased detection risk by security systems. Consequently, minimizing the number of queries remains a critical benchmark for evaluating the practicality and stealth of black-box attacks \cite{heu, sva}.
	
	Previous mentioned adversarial methods have  treated video data as a simple vector, disregarding its inherent multi-dimensional structure. In contrast, our paper introduces a novel approach that recognizes video data as a 4th-order tensor, fully leveraging its multi-dimensional multilinear properties. By considering the tensor representation of video data, our method can operate more effectively across different dimensions, offering increased flexibility and significantly reducing both the search space and the number of queries required—key advantages in the black-box setting.
	
	These improvements set our approach apart from existing techniques. The experimental results demonstrate the effectiveness of our method, showing superior attack success rates, reduced errors, and improved query efficiency compared to leading current methods. This innovative perspective not only advances the effectiveness of adversarial attacks but also pushes the boundaries of what can be achieved with efficient, low-query black-box attacks.
	The remainder of this paper is structured as follows: Section 2 covers the preliminaries and fundamental concepts. In Section 3, we provide a discussion of related work. Section 4 introduces the proposed methodology, while Section 5 presents the experimental results.
	
\section{Preliminaries}
	\label{Preliminaries} 
	An order-N tensor $\mathcal{A} \in \mathbb{R}^{I_1 \times \cdots \times I_N}$ is an n-dimensional array with elements denoted as $x_{i_1,\cdots,i_N}$, where $i_k = 1,\cdots,I_k$, $k = 1,\cdots,N$. 
	
	A slice is defined as a sub-tensor where only one index is fixed. For example, mode-1 slices of tensor $\mathcal{X} \in \mathbb{R}^{I_1 \times I_2 \times I_3 \times I_4}$ is denoted as:
	\begin{equation}
		x_{i_1,:,:,:}\in \mathbb{R}^{I_2 \times I_3 \times I_4}
	\end{equation}
	
	A fiber is defined as a sub-tensor where all indices except one are fixed. For example, a mode-1 fiber of tensor $\mathcal{X} \in \mathbb{R}^{I_1 \times I_2 \times I_3 \times I_4}$ is defined as follows:
	\begin{equation}
		x_{:,i_2,i_3,i_4} \in \mathbb{R}^{I_1}
	\end{equation}
Among the different tensor norms used for tensors, the well-known Frobenius norm is commonly applied. For a tensor $\mathcal{X} \in \mathbb{R}^{I_1 \times I_2 \times I_3 \times I_4}$ the Frobenius norm is defined as the square root of the sum of the squares of all its elements:
\begin{equation}
		||\mathcal{X}|| = \sqrt{\sum_{i_1}^{I_1} \sum_{i_2}^{I_2} \sum_{i_3}^{I_3} \sum_{i_4}^{I_4} x_{i_1 i_2 i_3 i_4}^2}
	\end{equation}
	
Let $u^{(i)} \in \mathbb{R}^{I_i}$, where $i=1,\dots,N$, represent first-order tensors (vectors). The outer product of these  vectors results in an order-N tensor $\mathcal{A} \in \mathbb{R}^{I_1 \times I_2 \times I_3 \times I_N}$, denoted as
\begin{equation}
	\mathcal{A}=u_1\circ u_2 \circ \cdots \circ u_N
\end{equation}
	with elements:
	\begin{equation}
		a_{i_1 i_2 \cdots, i_N} = u^{(1)}_{i_1} u^{(2)}_{i_2}\cdots u^{(N)}_{i_N}
\end{equation}

	The n-mode product of an M-order tensor $\mathcal{A} \in \mathbb{R}^{I_1 \times I_2 \times ... \times I_M}$ and a matrix $X \in \mathbb{R}^{K \times I_n}$ is defined as:
	\begin{equation}
		\mathbb{R}^{I_1 \times ... \times K \times I_n \times .. I_M} \ni \mathcal{B}  = \left( X \right)_n \cdot \mathcal{A}
	\end{equation}
	where,
	\begin{equation}
		b_{i_1,...,i_M} = \sum_{j=1}^{I_n} x_{i_n,j} a_{i_1,\cdots,i_{n-1}, j, i_{n+1}, \cdots,i_M}
	\end{equation}
	In alternative notation, this multiplication is denoted as $\mathcal{B} = \mathcal{A} \times_n X$.
\subsection{Tensor Decompositions and Corresponding Ranks}

Tensor decompositions, such as CANDECOMP/PARAFAC (CP) decomposition, Tucker or Higher-Order Singular Value Decomposition (HOSVD), as well as Tensor singular Value decomposition(TSVD) are well-known methods for analyzing tensors. These decompositions extend the Singular Value Decomposition (SVD) to higher-order tensors, providing valuable insights into multidimensional data and serving various roles in diverse applications.
For example, for an order-4 tensor $\mathcal{X} \in \mathbb{R}^{I_1 \times I_2 \times I_3 \times I_4}$, the CP decomposition approximates the tensor as:
\begin{equation}
	\label{cp1}
	\mathcal{X} \approx \sum_{i=1}^R \lambda_i \, u_i^{(1)} \circ u_i^{(2)} \circ u_i^{(3)} \circ u_i^{(4)}
\end{equation}

where $\lambda_i$ are scalar weights, and $u_i^{(j)} \in \mathbb{R}^{I_j}$ are factor vectors for each mode $j = 1, 2, 3, 4$. The symbol $\circ$ denotes the outer (tensor) product of vectors.

Each tensor decomposition introduces its own definition of tensor rank. In the CP decomposition, the tensor rank is defined as the smallest value of $R$ for which the approximation in equation~\eqref{cp1} becomes an exact equality:

\begin{equation}
	\mathcal{X} = \sum_{i=1}^R \lambda_i \, u_i^{(1)} \circ u_i^{(2)} \circ u_i^{(3)} \circ u_i^{(4)}
\end{equation}

That is, the CP rank of the tensor $\mathcal{X}$ is the minimal number of rank-1 tensors (outer products of vectors) needed to represent $\mathcal{X}$ exactly.

The $(r_1,\cdots, r_4)$-rank Tucker decomposition of $\mathcal{X}$ expresses the tensor as:
\begin{equation}
	\mathcal{X} \approx \sum_{i_1=1}^{r_1} \sum_{i_2=1}^{r_2} \sum_{i_3=1}^{r_3} \sum_{i_4=1}^{r_4} s_{i_1 i_2 i_3 i_4} \, u_{i_1}^{(1)} \circ u_{i_2}^{(2)} \circ u_{i_3}^{(3)} \circ u_{i_4}^{(4)}
\end{equation}
where $g_{i_1 i_2 i_3 i_4}$ are elements of the core tensor $\mathcal{S} \in \mathbb{R}^{r_1 \times r_2 \times r_3 \times r_4}$, and $u_{i_j}^{(j)} \in \mathbb{R}^{I_j}$ are the factor vectors for each mode. The integers $r_j$ represent the Tucker ranks along each mode. If we set $r_j = I_j$ for all $j$, the Tucker decomposition becomes the HOSVD:
\begin{equation}
	\mathcal{X} = \sum_{i_1=1}^{I_1} \sum_{i_2=1}^{I_2} \sum_{i_3=1}^{I_3} \sum_{i_4=1}^{I_4} s_{i_1 i_2 i_3 i_4} \, u_{i_1}^{(1)} \circ u_{i_2}^{(2)} \circ u_{i_3}^{(3)} \circ u_{i_4}^{(4)}
\end{equation}
Alternatively, this can be compactly written using tensor-matrix products:
\begin{equation}
	\mathcal{X} =\left(U^{(1)}, U^{(2)}, U^{(3)}, U^{(4)}\right)_{1:4} . \mathcal{S},
\end{equation}
where $\mathcal{S} \in \mathbb{R}^{I_1 \times I_2 \times I_3 \times I_4}$ is the core tensor, and $U^{(j)} = [u_1^{(j)}, \ldots, u_{I_j}^{(j)}] \in \mathbb{R}^{I_j \times I_j}$ are orthogonal matrices (singular matrices) corresponding to each mode $j$. In the HOSVD, the singular matrices $U^{(j)}$ are composed of orthogonal vectors, with the columns ordered according to the significance of their contribution to the data. The first columns of $U^{(j)}$ capture the most significant patterns (often smoother components), while the later columns capture finer details or noise. Therefore, the low-index columns $u_i^{(j)}$ play a crucial role in reconstructing the main features of the data, whereas higher-index columns represent minor variations.

The rank of a tensor in the Tucker or HOSVD decomposition is given by the multilinear rank, defined as $\operatorname{rank}(\mathcal{X}) = (r_1, r_2, r_3, r_4)$, where each $r_j$ is the dimension of the vector space spanned by the mode-$j$ fibers of the tensor. This means that $r_j$ is the rank of the mode-$j$ unfolding (matricization) of the tensor $\mathcal{X}$.	
These properties of the singular matrices in HOSVD have been effectively utilized in a wide range of applications. For instance, in image denoising and restoration, the leading singular vectors capture the most significant features of the data, enabling efficient removal of noise while preserving important structures \cite{Rezghi2017}. Similarly, in rs-fMRI classification, the use of HOSVD allows the extraction of crucial connectivity patterns in brain imaging, which leads to more accurate functional connectivity construction \cite{Noroozi2020}. HOSVD is also applied in dimensionality reduction, where the low-rank approximation helps in retaining the most important features while discarding noise and irrelevant details, making the data more manageable and enhancing computational efficiency. In all these applications, the properties of the singular matrices, especially their ability to order components by significance, play a key role in improving the effectiveness and efficiency of data reconstruction, noise reduction, feature extraction,  and classification.	
\section{Related Work and Background}

Let $x$ and $y$ be clean data and its true class, respectively. Also, let $f(\cdot)$ denote a Deep Neural Network (DNN). The goal of an adversarial attack is to produce a very slight and imperceptible perturbation, called an adversarial perturbation, that, when added to clean data $x$, fools the DNN into making a wrong prediction. This problem can be formulated as follows:

\begin{equation}\label{RW:e1}
	\begin{split}
		& \min ||\rho||_2^2 \\
		& \text{s.t. } f(x + \rho) \neq y.
	\end{split}
\end{equation}

The optimal solution of this problem, denoted by $\rho^{*}$, is the minimum imperceptible crafted adversarial perturbation that fools the DNN model $f(\cdot)$. 

Adversarial attacks on deep learning models have been an area of significant research in recent years, particularly due to their implications for the security and robustness of machine learning systems.  
Based on the level of access to network information, adversarial attacks are divided into two main categories: white-box and black-box methods.

In \textbf{white-box attacks}, the attacker has full knowledge of the target model, including its parameters, gradients, and structure, which allows them to craft perturbations effectively. Several prominent adversarial attack methods fall into this category. For example, the Fast Gradient Sign Method (FGSM) proposed by Goodfellow et al. \cite{Goodfellow2015} is a simple yet powerful attack that uses gradient information to generate adversarial examples quickly. Another widely studied attack is DeepFool, developed by Moosavi-Dezfooli et al. \cite{Moosavi2016}, which iteratively finds minimal perturbations to fool the model. These white-box methods illustrate the vulnerabilities of even well-trained models under complete transparency.

In contrast, \textbf{black-box attacks} assume no knowledge of the target model's internal structure, treating it as a black-box system. Attackers rely solely on querying the model by submitting inputs and observing outputs, which makes these attacks applicable to real-world scenarios where internal access is restricted. However, unlimited querying is often impractical or prohibited due to defense strategies that limit external queries to maintain security. Consequently, the number of queries becomes a crucial criterion when designing and assessing black-box attacks.
Techniques like transfer-based attacks \cite{Cheng2019} leverage the concept that adversarial examples crafted for one model can often transfer to successfully fool other models, exploiting the generality of adversarial vulnerabilities across architectures. Nonetheless, crafting successful adversarial examples in the black-box setting is exceptionally challenging due to the lack of information about the model. This challenge is further amplified when aiming to remain efficient and reduce computational costs.

In scenarios with hard-label outputs, the target function \( f \) is non-differentiable, and obtaining gradients explicitly is not feasible. Methods like the opt-attack \cite{opt} have been developed to address these challenges by effectively approximating the necessary information to generate adversarial examples. 
In the \textbf{opt-attack} method, the objective is to find an adversarial example by identifying the optimal perturbation direction $\rho$ that minimally alters the input $x$ to change the classifier's output. For each attack direction $\rho$, we define the function $g(\rho)$ as the minimal scalar $\lambda$ required to cause a misclassification when moving in the direction $\rho$:

\begin{equation}
\label{eq:g}
g(\rho) = \min_{\lambda} \left\{ \lambda \, \big| \, f\left( x + \lambda \frac{\rho}{\|\rho\|} \right) \neq f(x) \right\}, \quad \text{where} \quad y = f(x).
\end{equation}

This means that $g(\rho)$ represents the smallest perturbation magnitude along the direction $\rho$ that changes the model's prediction from $y$ to a different class. The optimal attack direction $\rho^{*}$ is the one that minimizes $g(\rho)$:

\[
g(\rho^{*}) = \min_{\rho} g(\rho).
\]

Once $\rho^{*}$ is determined, the adversarial example $x_{\text{adv}}$ is constructed as:

\[
x_{\text{adv}} = x + g(\rho^{*}) \frac{\rho^{*}}{\|\rho^{*}\|}.
\]

To solve this minimization problem in a black-box setting with hard-label outputs, the opt-attack \cite{opt} employs Zero-Order Optimization (ZOO) techniques to estimate the gradient of $g(\rho)$ with respect to $\rho$ without access to the model's internal parameters. Since $g(\rho)$ is non-differentiable, direct gradient computation is infeasible, and therefore the gradient is approximated using a directional finite difference along a randomly chosen direction. Specifically, the directional derivative of $g$ at $\rho$ in the direction of a random unit vector $u$ is approximated by:
\begin{equation} \label{eq:directional_derivative}
	\frac{\partial g(\rho)}{\partial \rho} \approx \frac{g\left( \rho + \beta u \right) - g\left( \rho \right)}{\beta} \, u,
\end{equation}

where $u$  drawn from a Gaussian distribution, and $\beta > 0$ is a small smoothing parameter.  This estimation of gradient, requires only two evaluations of $g(\rho)$ per iteration, making the opt-attack highly query-efficient, which is crucial in black-box settings where queries are often costly or limited.  Recent advancements over the opt-attack include methods like HopSkipJumpAttack (HSJA)\cite{Chen2020} and Sign-OPT \cite{Cheng2020}, which further improve query efficiency and reduce computational costs by employing more sophisticated gradient estimation techniques that converge faster and require fewer queries.

Recent research on adversarial examples has primarily focused on image classification models, exploring both white-box \cite{b:12,b:11,b:7,b:14,b:15} and black-box strategies \cite{b:16,b:17}. However, when it comes to video models, black-box adversarial attacks have received far less attention compared to their image model counterparts. This discrepancy arises from several key challenges.

First and foremost, videos represent high-dimensional data, incorporating not only spatial dimensions but also a temporal one, which makes them inherently more complex. While an image can be represented as $\mathcal{X} \in \mathbb{R}^{W \times H \times C}$—where $W$, $H$, and $C$ denote the width, height, and number of channels respectively, a video adds a temporal dimension, represented as $\mathcal{X} \in \mathbb{R}^{W \times H \times C \times T}$, where $T$ denotes time. This added complexity makes designing adversarial attacks for videos significantly more challenging than for images.

Adapting adversarial attacks developed for image models to video models involves substantial challenges in terms of time, computational cost, and the number of queries required—often making such attacks easily detectable. Some of the earlier black-box adversarial attacks on video models, such as VBAD \cite{vbad}, simply extended image-based adversarial methods to entire video sequences. However, attacking the entire video not only results in a high query cost but also leads to inefficiencies and low-quality adversarial examples. As a result, there is a growing need to develop black-box attacks that are tailored specifically for video models. Recent studies have taken steps to reduce the inherent complexities of attacking video data by introducing a key frame selection approach. This approach involves identifying and targeting only a subset of frames that are more significant for the classification task, rather than perturbing every frame. By doing so, attackers can focus on these key frames, while leaving the non-key frames unchanged. While this strategy is intended to improve the efficiency of adversarial attacks by reducing the number of frames that need to be perturbed, it introduces a new challenge: the selection of key frames. Finding the optimal subset of key frames is critical, as it directly impacts the success of the attack. Unfortunately, this selection process itself can increase the number of queries required, thereby leading to inefficiencies.

In general, key-frame-based methods represent the video $\mathcal{X}$ as $\mathcal{X} = (\mathcal{X}^n, \mathcal{X}^k)$, where $\mathcal{X}^k$ denotes the key frames, and $\mathcal{X}^n$ represents the non-key frames. The overarching goal is to determine the most effective key frame subset and subsequently generate an optimal adversarial example based on it. As an example, this problem in the Heuristic attack \cite{heu}, which is a black-box key frame-based attack on video recognition models, can be formulated as follows:

\begin{equation} \label{RW:e2}
	\begin{split}
		& \min_{\mathcal{X}^k} \min_{\rho} g(\rho) \\
		& g(\rho) = \min_{\lambda} \left( f(\mathcal{X}^n, \mathcal{X}^k + \lambda \rho ) \neq f(\mathcal{X}^n, \mathcal{X}^k) \right)
	\end{split}
\end{equation}

Selecting $k$ frames out of $n$ frames is inherently challenging, as the number of possible combinations is $\binom{n}{k}$. The approach proposed in \cite{heu} addresses this by selecting key frames through a heuristic method to find an effective key frame set based on each frame's role and significance in the classification task. The selection process involves removing each frame individually, evaluating the video classification without that frame, and calculating an adversarial score to determine its impact on the classification. Frames are then ranked by their adversarial score, and those with the highest scores are selected as key frames for each video.

Despite its success in fooling video models, the Heuristic attack still requires a considerable number of queries, which limits its practical applicability. Moreover, while the key frame selection is based on each frame's discriminatory significance, the method lacks a holistic evaluation of different possible subsets since it only assesses frames individually. Consequently, the solution obtained does not effectively solve equation \ref{RW:e2} in an optimal manner, as it overlooks the potential interactions between frames that could be critical for a truly optimal attack.

Additionally, the SVA attack \cite{sva} introduces an innovative approach to key frame selection by leveraging reinforcement learning. However, it still requires numerous queries to the black-box model in order to determine the optimal subset of key frames. Sparse adversarial attacks, such as Heuristic \cite{heu} and SVA \cite{sva}, which focus on selecting and perturbing only key frames, demonstrate superior performance compared to non-sparse methods like VBAD \cite{vbad}, which attempt to perturb all frames indiscriminately. Despite their improved efficiency and targeted approach, these sparse attacks still face several limitations:

\begin{figure}
	\centering
	\includegraphics[width=13cm]{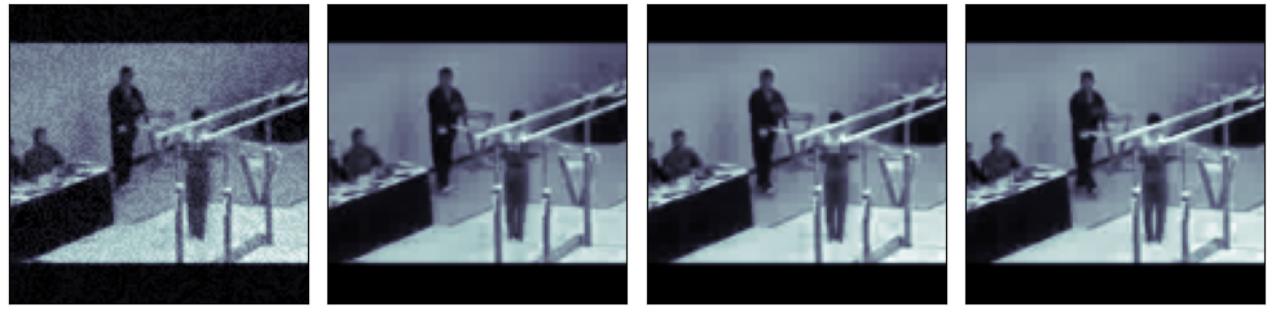}
	\caption{The figure indicates the frames of an adversarial example generated by SVA \cite{sva}. The first frame denotes an adversarial frame (keyframe) while others (non-keyframes) have remained unchanged. The difference between a keyframe and other frames is visually detectable.}
	\label{fig:3}
\end{figure}
\begin{itemize}
 \item The core approach of both Heuristic and SVA attacks \cite{heu,sva} is to select a subset of keyframes and perturb only those frames, leaving the others untouched. However, as shown in Figure \ref{fig:3}, the distinction between keyframes and non-keyframes can be visually apparent, especially when examining the entire video. While perturbations may seem subtle when viewing keyframes individually, the overall difference between perturbed keyframes and unaltered frames becomes noticeable, which diminishes the effectiveness of the attack. Therefore, selecting the most suitable keyframes is critical. A significant limitation of Heuristic \cite{heu}, SVA \cite{sva}, and similar approaches is their failure to consider the interactions among all frames in a video, which prevents a holistic optimization and can lead to suboptimal adversarial examples.
 
 \item Although selecting and attacking only keyframes reduces the search space for generating an adversarial video from an order of $O(W \times H \times C \times T)$ to $O(W \times H \times C \times K)$ (where $K$ is the number of keyframes), these methods still require a considerable number of queries and significant computational time. This is because, even with fewer frames, the spatial dimensions $W \times H$ contribute heavily to the overall complexity of the search space. Therefore, despite recent methods \cite{heu,sva} aiming to increase attack efficiency by selecting keyframes, the search space remains expansive, with a large number of parameters, resulting in high query demands and added complexity.
 
 \item In addition to these challenges, all existing adversarial video attacks, including those that operate on keyframes, do not fully utilize the multi-dimensional structure of video data and instead treat it as a flattened vector. However, multi-dimensional data like video and image sequences have unique properties and should be analyzed using specialized tools. For example, by considering a video as a tensor, we can independently analyze dimensions corresponding to rows, columns, colors, and time frames, which provides far more nuanced information than treating the video as a one-dimensional vector. Viewing the video through a multi-dimensional tensor perspective could thus offer significant advantages, as demonstrated by numerous recent applications in the field.
 \end{itemize}

In the next section, we propose a novel tensor-based approach to generate black-box adversarial attacks for video recognition models, leveraging the unique characteristics of video data. By representing video as a four-dimensional tensor, with each dimension corresponding to a distinct aspect of the data, such as spatial and temporal components, we address the key limitations of previous methods. Our approach effectively generates successful, imperceptible adversarial perturbations while achieving a significant reduction in the number of queries, search space, and overall computational complexity.

\section{A Novel Tensor Based Low-rank Black-Box Attack for Video Classification}

In this section, we present a novel tensor-based low-rank adversarial attack approach for video data. Let $\mathcal{X} \in \mathbb{R}^{W \times H \times C \times T}$ represent a clean video tensor, where $W$, $H$, $C$, and $T$ denote the width, height, number of color channels, and number of frames (time steps), respectively. The video is correctly classified by a classifier $f$ with the true label $y$, i.e., $f(\mathcal{X}) = y$.

The goal of an adversarial attack is to find a perturbed video $\mathcal{X}_{\text{adv}}$ such that the classifier misclassifies it, i.e., $f(\mathcal{X}_{\text{adv}}) \ne y$, while keeping the perturbation imperceptible or within certain constraints. This can be formulated as:

\begin{equation}
	\label{eq:adv_perturbation}
	\mathcal{X}_{\text{adv}} = \mathcal{X} + \mathcal{E},
\end{equation}
where $\mathcal{E} \in \mathbb{R}^{W \times H \times C \times T}$ represents the adversarial perturbation tensor.

Existing adversarial attack methods typically consider the vectorized form of the data, treating the multi-dimensional tensor as a flat vector:
\begin{equation}
	x_{\text{adv}} = x + e,
\end{equation}
where
\begin{equation}
	\label{eq:vectorization}
	x_{\text{adv}} = \operatorname{Vec}(\mathcal{X}_{\text{adv}}), \quad
	x = \operatorname{Vec}(\mathcal{X}), \quad
	e = \operatorname{Vec}(\mathcal{E}),
\end{equation}
and $\operatorname{Vec}: \mathbb{R}^{W \times H \times C \times T} \rightarrow \mathbb{R}^{WHCT}$ denotes the vectorization operator that reshapes the tensor into a vector of length $WHCT$.
This vectorization approach ignores the inherent multi-dimensional structure of video data and the unique characteristics of each mode(spatial dimensions, color channels, and temporal dimension). As a result, existing methods not only suffer from a large search space of size $O(W H C T)$ when seeking adversarial perturbations(leading to high computational complexity), but also miss the opportunity to exploit the data's inherent low-rank properties for more efficient attacks by disregarding the correlations and structures within the tensor modes.

\begin{figure}
		\centering
		\includegraphics[width=6cm]{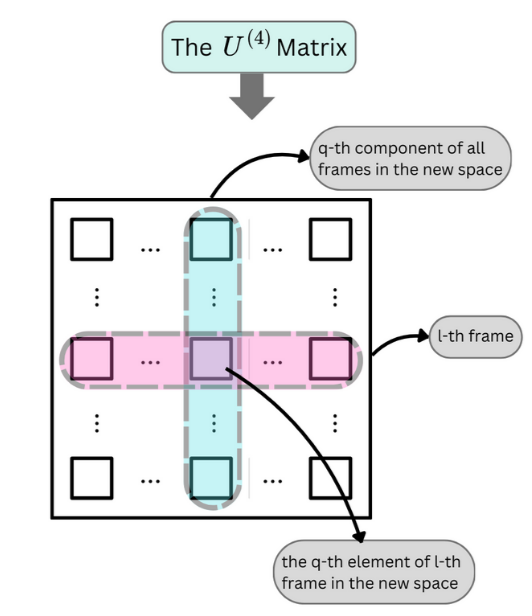}
		\caption{After Final Review will be added}
		\label{Tensors}
	\end{figure}

\begin{figure}
	\centering
	\includegraphics[width=12cm]{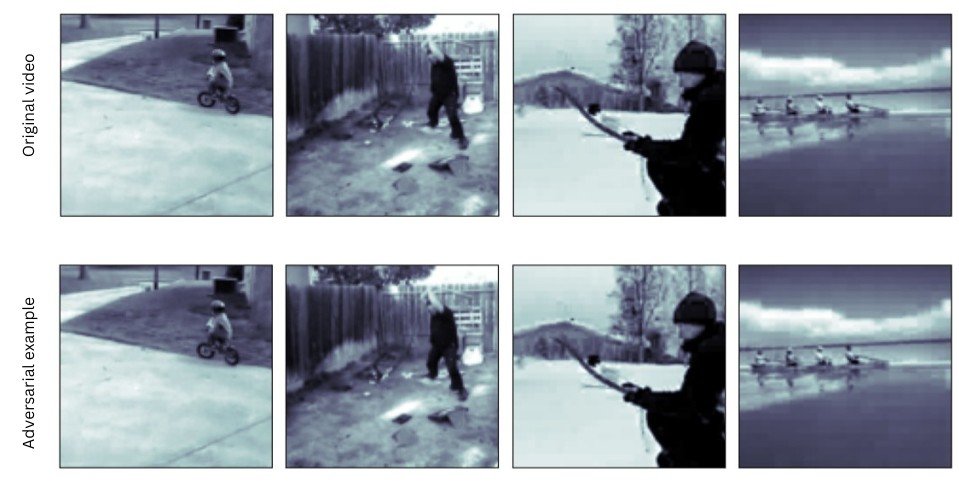}
	\caption{Generating imperceptible rank-1 adversarial perturbations on 4-frame videos. The images above are the clean frames, while the images below are adversarial frames generated by introducing rank-1 perturbations.}
	\label{fig:f1}
\end{figure}

Leveraging the multidimensional structure of video data allows for a distinct treatment of each mode, providing deeper analytical insights. Multi-dimensional tensor decompositions, such as CANDECOMP/PARAFAC (CP) and Tucker, are highly effective tools that extract latent structures and correlations from complex datasets. By breaking down a tensor into its fundamental components, these methods reveal the underlying patterns and independencies across different modes, allowing us to identify latent factors that can be effectively targeted for further analysis or intervention. The tensor rank, derived from these decompositions, serves as an indicator of data complexity and information content, playing a crucial role in identifying the optimal representation of the data, thus guiding effective dimensionality reduction and interpretation.	
To provide more insight, consider the CP decomposition of the video tensor $\mathcal{X} \in \mathbb{R}^{W \times H \times C \times T}$ with an appropriate rank $r$ as:	
\begin{equation}
		\label{eq:cp_decomposition1}
		\mathcal{X} \approx \sum_{k=1}^{r} \lambda_k \, u_k^{(1)} \circ u_k^{(2)} \circ u_k^{(3)} \circ u_k^{(4)},
\end{equation}
where $\lambda_k \in \mathbb{R}$ are scalar weights, and $u_k^{(j)} \in \mathbb{R}^{I_j}$ are factor vectors for mode $j$ (with $I_1 = W$, $I_2 = H$, $I_3 = C$, $I_4 = T$). The matrices $U^{(j)} = [u_1^{(j)}, u_2^{(j)}, \ldots, u_r^{(j)}] \in \mathbb{R}^{I_j \times r}$ for $j = 1, 2, 3, 4$ capture the latent components across each mode of the tensor.
To demonstrate this, consider the fourth dimension. Here, by using \eqref{eq:cp_decomposition}, we can express the $l$-th frame of the video, i.e., $\mathcal{X}_{:,:,:,l}$, by fixing the temporal index to $i_4 = l$ as follows:	
	\begin{equation}
		\label{eq:cp_frame}
		\mathcal{X}_{:,:,:,l} \approx \sum_{k=1}^{r} u_{lk}^{(4)} \lambda_k \, u_k^{(1)} \circ u_k^{(2)} \circ u_k^{(3)} = \sum_{k=1}^{r} u_{lk}^{(4)} \, \mathcal{A}_k = \left(u^{(4)}_{l,:}\right)_4 . \mathcal{A}
	\end{equation}
where the order-4 tensor $\mathcal{A}$, with slices $\mathcal{A}_{:,:,:,k}:=\mathcal{A}_k= \lambda_k \, u_k^{(1)} \circ u_k^{(2)} \circ u_k^{(3)} \in \mathbb{R}^{W \times H \times C}$  capturing spatial and color information. Here, $u_{lk}^{(4)}$ is the $l$-th element of the factor vector $u_k^{(4)}$ corresponding to the temporal mode, and $u^{(4)}_{l,:}$ is the $l$-th row of $U^{(4)}$.
This formulation reveals that each frame can be represented as a linear combination of a set of rank-one spatial tensors $\mathcal{A}_k$, weighted by the temporal factors $u_{lk}^{(4)}$. Thus, the $l$-th row of $U^{(4)}$ represents the $l$-th frame in the new latent space. In the same way one can see that  for each $q$, $q$-th columns of $U^{(4)}$, shows the effect of  $\mathcal{A}_q$ is construction of all frames. Figure \ref{Tensors}, show this schematically. 	The same results can also be derived for other modes.
Similarly, the Tucker decomposition or Higher-Order Singular Value Decomposition (HOSVD) can be employed to capture the multi-mode structure of the tensor. The $(r_1, r_2, r_3, r_4)$-rank Tucker decomposition of $\mathcal{X}$ can be expressed as:
	\begin{equation}\label{eq:hosvd}
		\mathcal{X} \approx \left(U^{(1)}, U^{(2)}, U^{(3)}, U^{(4)}\right)_{1:4} . \mathcal{S},
	\end{equation}
which allows us to approximate the tensor as a sum of rank-one components:
	\begin{equation}
		\label{eq:hosvd_sum}
		\mathcal{X} \approx \sum_{i_1=1}^{r_1} \sum_{i_2=1}^{r_2} \sum_{i_3=1}^{r_3} \sum_{i_4=1}^{r_4} s_{i_1 i_2 i_3 i_4} \, u_{i_1}^{(1)} \circ u_{i_2}^{(2)} \circ u_{i_3}^{(3)} \circ u_{i_4}^{(4)}.
	\end{equation}
By fixing the temporal index $i_4 = l$, the $l$-th frame can be expressed as:
	\begin{equation}
		\label{eq:hosvd_frame}
		\mathcal{X}_{:,:,:,l} \approx (U^{(4)}_{l,:})_4 . \Tilde{\mathcal{S}},
	\end{equation}
where $\mathbb{R}^{I_1 \times I_2 \times I_3 \times r_4} \ni \Tilde{\mathcal{S}} = \left(U^{(1)}, U^{(2)}, U^{(3)}, U^{(4)}\right)_{1:3} . \mathcal{S}$. Thus, the $l$-th row of $U^{(4)}$ is the representation of the $l$-th frame in the new latent space. Similar conclusions can also be drawn for other modes.
These decomposition demonstrate that the tensor $\mathcal{X}$ is represented as a summation of rank-one tensors. In both cases, the matrices $U^{(j)}, j=1,2,3,4$, capture the latent features corresponding to each mode of the video tensor, specifically, the spatial rows, columns, color channels, and temporal frames.  
In the following sections, we detail the specific algorithm  used to implement our tensor-based attacks, demonstrating their superiority over existing  approaches in both performance and computational efficiency.
Based on the above analysis, we found that each tensor can be represented as the summation of rank-one tensors, which each rank-one tensor consists of the outer product of vectors corresponding to different modes of the tensor. With this viewpoint, the perturbation $\mathcal{E}$ can be considered as a low-rank tensor (e.g., rank-1):
\begin{align*}
	\mathcal{E} = \sum_{i=1}^{l}\theta_i^{(1)} \circ \theta_i^{(2)} \circ \theta_i^{(3)} \circ \theta_i^{(4)}
\end{align*}
This representation gives the ability to separately handle different features of the tensor. So, the proposed model based on low rank attack can be formulated as follows:
\begin{equation}
	\label{man1}
	\begin{split}
		& \min \mathcal{L} (\mathcal{E}) \\
		& \text{s.t. } \begin{cases}
			f(\mathcal{X} + \mathcal{E}) \neq f(\mathcal{X}) \\
			\mathcal{E} = \sum_{i=1}^{l}\theta_i^{(1)} \circ \theta_i^{(2)} \circ \theta_i^{(3)} \circ \theta_i^{(4)}, \quad \theta_i^{(j)} \in \mathbb{R}^{I_j} 
		\end{cases} 
	\end{split}
\end{equation}
Here, the loss function $\mathcal{L}(.)$ can be defined in two ways:
\begin{equation} \label{mq8}
	\mathcal{L}(\mathcal{E}) = ||\mathcal{E}||_F^2 =\sum_{i=1}^{l} \prod_{j=1}^{4} ||\theta_i^{(j)}||_2^2
\end{equation}
or 
\begin{equation} \label{mq9}
	\mathcal{L} (\mathcal{E}) = \sum_{i=1}^{l}\sum_{j=1}^{4} ||\theta_i^{(j)}||_2^2
\end{equation}
Equation \ref{mq8} measures the overall perturbation related to the video, while Equation \ref{mq9} represents the sum of perturbations across all modes. In this paper, to compare our attack to other attacks, we use Equation \ref{mq8}. From our practical experience, considering rank-1 perturbation (i.e., $l=1$) is enough to perform a successful attack.
In fact in Equation \eqref{man1}, we imposed a low-rank constraint on the error term, while controlling the magnitude of this error through the loss function. Although a low-rank constraint does not necessarily guarantee minimal perturbation when evaluated using norm-based metrics, there are instances where a low-rank error introduces significant perturbation according to these metrics, yet the visual difference remains imperceptible. This observation demonstrates the limitations of norm-based evaluations in capturing perceptual similarity.
For example, we constructed a rank-1 tensor $\mathcal{I} \in \mathbb{R}^{W \times H \times C \times T}$, where all elements were equal to one. We then scaled it by a large value, such as 256, which did not affect its rank, which remained one. This rank-1 perturbation was added to clean frames to generate a rank-1 adversarial example,  as some shown in Figure \ref{fig:f1}.
The resulting adversarial examples led to incorrect predictions by the classifier, despite the adversarial frames being visually similar to the clean frames, as shown in Figure \ref{fig:f1}. The mean absolute perturbation metric indicates a significant difference, suggesting that the clean and perturbed frames should be distinct. However, the rank-1 adversarial examples still exhibit an almost indistinguishable similarity to the clean frames. This highlights that vector-based metrics like mean absolute perturbation are insufficient for evaluating adversarial examples in video or image contexts effectively.

This new approach provides several key advantages. Firstly, by assigning distinct values to each parameter $\theta_i$, the attacker can effectively target specific features and dimensions of the video $\mathcal{X}$ in their own respective spaces. For example, $\theta_1$ focuses exclusively on the frames within its designated space.

Additionally, by representing the perturbation $\mathcal{E}$ as a summation of low-rank components, the search space required for determining $\mathcal{E}$ is significantly reduced. If

\begin{equation*}
	\begin{split}
		 \varphi = \{ \mathcal{E} \mid \mathcal{E} \in \mathbb{R}^{T \times C \times W \times H} \}, \quad 
		 \psi = \{ \mathcal{E} \mid \mathcal{E} = \sum_{i=1}^{l} \theta^{(1)}_i \circ \theta^{(2)}_i \circ \theta^{(3)}_i\circ \theta^{(4)}_{i} \}
	\end{split}
\end{equation*}
represent the search space of the classical method and our proposed low-rank attack, respectively, it is evident that $\psi \subset \varphi$, indicating a substantial reduction in the search space. As a result, the number of parameters in the classical approach for video $\mathcal{X} \in \mathbb{R}^{W \times H \times C \times T}$ is of order $O(TCWH)$, whereas in the proposed method, it is reduced to $O(T + C + W + H)$. This reduction considerably improves the efficiency of the attack by requiring fewer queries.
\subsection{Solving the Proposed Model for Black-Box Adversarial Attacks}

In this section, we solve the model the proposed tensor-based low-rank attack, named the \textbf{TenAd} Attack, which operates within a Black-Box setting under Hard labeling constraints. This solver builds upon the foundation of the opt-attack \cite{opt}, but is specifically adapted to leverage low-rank modeling for enhanced efficiency and effectiveness. Our objective is two-fold: first, to determine the minimal distance required in each direction to induce misclassification, and second, to identify the optimal direction that yields an adversarial example. For simplicity, and without loss of generality, we focus on the case where $l = 1$.
We define the perturbation as follows:
\begin{equation}
	\mathcal{E} = \lambda \frac{\theta^{(1)}}{\|\theta^{(1)}\|} \circ \frac{\theta^{(2)}}{\|\theta^{(2)}\|} \circ \frac{\theta^{(3)}}{\|\theta^{(3)}\|} \circ \frac{\theta^{(4)}}{\|\theta^{(4)}\|},
\end{equation}
where $\theta^{(j)}, j = 1, \cdots, 4$  are the perturbation directions in each mode, and $\lambda$ represents the magnitude of the perturbation. The loss associated with this perturbation is given by $\mathcal{L} = \prod_{j=1}^{4} \|\theta^{(j)}\|_2 = \lambda$. Thus, for each direction, the optimal step length can be computed as:

\begin{equation}
	\begin{split}
		& g(\theta) = \min_{\lambda} \left(f\left(\mathcal{X} + \lambda \frac{\theta^{(1)}}{\|\theta^{(1)}\|} \circ \frac{\theta^{(2)}}{\|\theta^{(2)}\|} \circ \frac{\theta^{(3)}}{\|\theta^{(3)}\|} \circ \frac{\theta^{(4)}}{\|\theta^{(4)}\|} \right) \neq y \right),
	\end{split}
\end{equation}
where $y = f(\mathcal{X})$ denotes the true label of the input $\mathcal{X}$.
The adversarial direction is represented by a set of low-rank components: specifically, $\theta^{(1)}$, $\theta^{(2)}$, $\theta^{(3)}$, and $\theta^{(4)}$. These components collectively form the parameter vector $\theta = [\theta^{(1)\sf T}, \cdots, \theta^{(4)\sf T}]^{\sf T}$. Correspondingly, the distance function $g(\theta)$ quantifies the extent of perturbation required to induce misclassification. So, the optimal parameters $\theta^*$ that minimizes the required perturbation distance, can be expressed as:
\begin{equation}
	\begin{split}
		& g(\theta^* ) = \min_{\theta} g(\theta)
	\end{split}
\end{equation}
The final adversarial example $\mathcal{X}_{adv}$ is then generated by applying the optimal perturbation, resulting in:
\begin{equation}
	\mathcal{X}_{adv} = \mathcal{X} + g(\theta^*) \left( \frac{\theta^{(1)*}}{\|\theta^{(1)*}\|} \circ \frac{\theta^{(2)*}}{\|\theta^{(2)*}\|} \circ \frac{\theta^{(3)*}}{\|\theta^{(3)*}\|} \circ \frac{\theta^{(4)*}}{\|\theta^{(4)*}\|} \right)
\end{equation}

In summary, this solver for the \textbf{TenAd} Attack seeks to efficiently determine both the optimal direction and the corresponding minimal perturbation required to craft an adversarial example using a low-rank representation. By leveraging the reduced dimensionality of low-rank components, our method aims to achieve effective adversarial attacks with minimal computational overhead, even in the challenging Black-Box, Hard labeling setting.

In order to determine $g(\theta^*)$, we follow a similar approach to the opt-attack \cite{opt}. Like the opt-attack, we utilize gradient descent optimization. However, since our attack operates within a black-box setting, we employ zero-order optimization \cite{b:16} to estimate the true gradient:

\begin{equation}
	\frac{\partial g}{\partial \theta} = \begin{pmatrix}
		\frac{\partial g}{\partial \theta^{(1)}} \\
		\vdots \\
		\frac{\partial g}{\partial \theta^{(4)}}
	\end{pmatrix}, \quad \frac{\partial g}{\partial \theta_i} \in \mathbb{R}^{I_i}.
\end{equation}
where the partial derivative for each component is given by:
\begin{equation}
	\frac{\partial g}{\partial \theta^{(j)}} = \frac{g(\overline{\theta}_{-j}) - g(\theta)}{\beta} \cdot u_j
\end{equation}
and $\overline{\theta}_{-j} = [\overline{\theta}^{(1)\sf T}, \cdots, \overline{\theta}^{(4)\sf T}]^{\sf T}$, with:

\begin{align*}
	\overline{\theta}^{(i)} =
	\begin{cases}
		\theta^{(i)}, & \quad i \neq j \\
		\theta^{(j)} + \beta u_{j}, & \quad i = j
	\end{cases}
\end{align*}
Here, $u_j$ represents a random Gaussian vector, while $\beta > 0$ serves as a smoothing parameter, which is subject to reduction by a factor of ten if the estimated gradients do not provide meaningful insights. To update the components of $\theta$ using gradient descent, we proceed as follows:

\begin{equation}
	\theta^{\text{NEW}} = \theta^{\text{OLD}} - \alpha \cdot \frac{\partial g}{\partial \theta}
\end{equation}
where $\alpha$ is the learning rate.

An alternative way to estimate the gradient is to define $\rho =  \theta^{(1)}\circ \theta^{(2)}\circ \theta^{(3)}\circ \theta^{(4)}$. In this approach,  $g(\rho)$ becomes as \eqref{eq:g} and all derivatives are computed using the chain rule, starting from $\frac{\partial g}{ \partial \rho}$. For instance:

\begin{equation} \label{eq45}
	\begin{split}
		\frac{\partial g}{\partial \theta^{(1)}_j} 
		 = \sum_{i_1, i_2, i_3, i_4} 
		\frac{\partial g}{ \partial \rho_{i_1, i_2, i_3, i_4}}
		\frac{\partial \rho_{i_1, i_2, i_3, i_4}}{\partial \theta^{(1)}_j} 
		 = \sum_{i_2, i_3, i_4} \frac{\partial g}{ \partial \rho_{j, i_2, i_3, i_4}}
		\theta^{(2)}_{i_2} \theta^{(3)}_{i_3} \theta^{(4)}_{i_4}
	\end{split}
\end{equation}
By the definition of the matrix-tensor product, this can be expressed as:
\begin{equation}
	\frac{\partial g}{\partial \theta^{(1)}} = \left(\theta^{(2)}, \theta^{(3)}, \theta^{(4)}\right)_{2:4} . \left( \frac{\partial g}{ \partial \rho} \right)
\end{equation}
The same relation holds for the other $\theta^{(j)}$ components:
\begin{align*}
	\frac{\partial g}{\partial \theta^{(2)}} = \left(\theta^{(1)}, \theta^{(3)}, \theta^{(4)}\right)_{1,3,4} . \left( \frac{\partial g}{ \partial \rho} \right) \\
	\frac{\partial g}{\partial \theta^{(3)}} = \left(\theta^{(1)}, \theta^{(2)}, \theta^{(4)}\right)_{1,2,4} . \left( \frac{\partial g}{ \partial \rho} \right) \\
	\frac{\partial g}{\partial \theta^{(4)}} = \left(\theta^{(1)}, \theta^{(2)}, \theta^{(3)}\right)_{1:3} . \left( \frac{\partial g}{ \partial \rho} \right)
\end{align*}
The gradient $\frac{\partial g}{ \partial \rho}$ is approximated as follows:
\begin{equation}
	\frac{\partial g}{ \partial \rho} = \frac{g(\rho + u \cdot \beta) - g(\rho)}{\beta} \cdot \mathcal{U}
\end{equation}
where $\mathcal{U}$ is random tensor from Gaussian distribution.
By exploiting the properties of the matrices \( U^{(j)} \), in both CP and Tucker decomposition, we can effectively initialize the vectors \( \theta^{(j)} \) using the columns of these matrices. specially in Tucker in setting \( \theta^{(j)} = u_q^{(j)} \) with a small index $q$ enables the attack to focus on the dominant or general characteristics of the \( j \)-th attribute. Conversely, increasing the value of \( q \) allows the attack to target more intricate details of that attribute, effectively controlling the granularity of the perturbation.
\section{Experiments}
This section presents a comprehensive evaluation aimed at assessing the performance of our novel adversarial attack, termed TenAd. To this end, we compared our black-box attack with state-of-the-art methods in attacking video classification, such as Heuristic~\cite{heu}, SVA~\cite{sva}, and VBAD~\cite{vbad}.

Our evaluation utilizes two widely employed video datasets: UCF-101~\cite{ucf} and HMDB-51~\cite{hmdb}. UCF-101, sourced from YouTube, comprises 13,320 videos spanning 101 action classes. Meanwhile, HMDB-51, a substantial human motion dataset, includes 7,000 videos belonging to 51 action categories. Following the procedure detailed in~\cite{heu}, we extracted 16-frame from each video using uniform sampling.
We selected a video recognition model, namely C3D~\cite{ucf}, as our target model. In this model, we assume that the attacker's access is limited to the top-1 predicted class. While our approach primarily focuses on untargeted attacks, it is important to note that our framework can be extended to encompass targeted attacks.
Also, we employ  the following  metrics to comprehensively comparing the quality of the methods:
\begin{itemize}
\item \textbf{Mean Query Number (MQ):} This signifies the average number of queries necessary to generate each adversarial example.

\item \textbf{Mean Absolute Perturbation (MAP):} This indicates the average perturbation across all  videos:
\begin{equation}
	\text{MAP} = \frac{1}{N} \sum_{i=1}^{N} \frac{\|\mathcal{X}_{i,\text{adv}} - \mathcal{X}_i\|_1}{m}
\end{equation}
where $m=WHCT$, where $W$, $H$, $C$, and $T$ represent the width, height, number of channels, and number of frames of the video, respectively and $N$ denotes the number of tests.

\item \textbf{Mean Structural Similarity Index Measure (MSSIM):} SSIM is a measure used to capture the similarity between images. For videos, MSSIM is computed by averaging the SSIM of all frames across all videos.
\end{itemize}
All the mentioned criteria were originally developed for images. However, some methods, such as those that attack only specific frames or regions, may introduce a large amount of perturbation on a few frames, which is not a favorable scenario. This is because we compute the evaluation criteria over all pixels of the video. Therefore, to focus on the perturbed frames and regions, we introduce the \textbf{MAP*} and \textbf{SSIM*} criteria, which are adapt \textbf{MAP} and \textbf{SSIM} works only on the perturbed frames and regions.
\begin{table}[h!]
	\centering
	\begin{tabular}{l|c|c}
		\textbf{Attack} & \textbf{UCF101} & \textbf{HMDB51} \\
		\hline 
		TenAd & 243 & 243  \\
		Heuristic~\cite{heu} & 75,264 & 150,528 \\
		SVA~\cite{sva} & 263,424 & 225,792 \\
		VBAD~\cite{vbad} &  602,112 & 602,112 \\
	\end{tabular}
	\caption{Number of parameters per test sample for each attack on the C3D model using UCF101 and HMDB51 datasets.}
	\label{table3}
\end{table}
It is evident that the proposed method, TenAd, requires significantly fewer parameters compared to the other methods.
Table~\ref{table1} presents a comparison of the quality of the proposed method, TenAd, with other state-of-the-art methods based on mentioned criteria.

\begin{table}[h!]
	\centering
	\begin{adjustbox}{width=\columnwidth,center}
		\begin{tabular}{ll|cccccccc}
			\hline
			\textbf{Model} & \textbf{Dataset} & \textbf{Attack} & \textbf{MQ} & \textbf{MT} & \textbf{MAP} & \textbf{MAP*} & \textbf{SSIM} & \textbf{SSIM*} & \textbf{FR} (\%) \\
			\hline\hline
			\multirow{4}{*}{C3D} 
			& HMDB & TenAd & \textbf{22.36} & 6.60 & 73.41 & \textbf{4.59} & \textbf{0.391} & \textbf{0.024} & \textbf{99.21} \\
			& HMDB & Heuristic~\cite{heu} & 5,947.9 & 11.05 & 94.32 & 7.38 & 0.101 & $6.3 \times 10^{-3}$ & 99.16 \\   
			& HMDB & SVA~\cite{sva} & 3,328.9 & \textbf{4.67} & \textbf{56.84} & 9.99 & $1.7 \times 10^{-4}$ & $2.9 \times 10^{-5}$ & 96.33 \\ 
			& HMDB & VBAD~\cite{vbad} & 68,584.2 & 32.24 & 59.59 & \textbf{3.72} & $7.6 \times 10^{-5}$ & $2.1 \times 10^{-5}$ & 95.00 \\  
			\hline
			\multirow{4}{*}{C3D} 
			& UCF & TenAd & \textbf{22.47} & \textbf{6.84} & 68.64 & \textbf{4.29} & \textbf{0.420} & \textbf{0.0262} & \textbf{98.10} \\ 
			& UCF & Heuristic~\cite{heu} & 53,596.4 & 55.54 & 96.15 & 6.01 & 0.022 & 0.0014 & 97.50 \\   
			& UCF & SVA~\cite{sva} & 4,473.8 & 7.11 & \textbf{53.24} & 10.35 & $1.2 \times 10^{-4}$ & $2 \times 10^{-5}$ & 86.90 \\   
			& UCF & VBAD~\cite{vbad} & 71,480.8 & 31.68 & 56.50 & \textbf{3.53} & $5.2 \times 10^{-5}$ & --- & 87.00 \\  
			\hline
		\end{tabular}
	\end{adjustbox}
	\caption{Comparison of our attack with state-of-the-art methods on the C3D model using HMDB51 and UCF101 datasets.}
	\label{table1}
\end{table}

As shown in Table~\ref{table1}, our method, TenAd, consistently outperforms other techniques in terms of the fooling rate (FR) on both datasets. This is achieved with significantly fewer queries (MQ), despite our approach targeting the entire video rather than just a subset of keyframes. Notably, TenAd surpasses VBAD~\cite{vbad} across multiple metrics, including MQ, SSIM, SSIM*, and FR on both datasets.

Furthermore, although our method exhibits a higher overall mean absolute perturbation (MAP) compared to SVA~\cite{sva}, it achieves lower per-frame perturbation (MAP*)—a critical factor in human perceptibility—compared to SVA~\cite{sva} and Heuristic~\cite{heu}, making it visually more imperceptible than these methods. While VBAD~\cite{vbad} achieves lower MAP and MAP* on both datasets, TenAd still surpasses it in terms of visual imperceptibility, as indicated by higher SSIM and SSIM* values.

Moreover, our method consistently generates adversarial examples with substantially higher SSIM similarity to the original videos compared to other methods, underscoring TenAd's capability to produce visually similar adversarial examples. This means that despite not selecting a subset of keyframes, our attack achieves closer visual resemblance to the original video. This is due to our incorporation of smaller adversarial perturbations in each frame, distinguishing our method from sparse attack strategies.

Notably, our approach achieves higher success rates (FR), requires fewer queries (MQ), and generates adversarial examples that closely resemble the original videos.

Figure \ref{fig:comp} demonstrates some perturbed frames generated by our method. Here the low rank property of attacks could be seen.

\begin{figure*}[h!]
	\centering
	\includegraphics[width=0.8\textwidth]{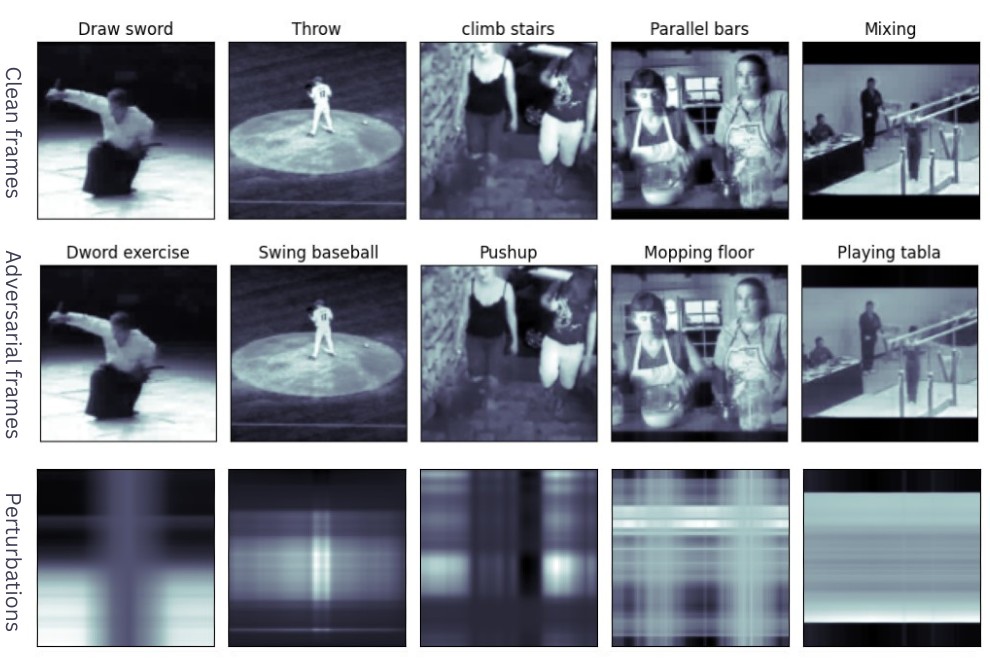}
	\caption{Low rank perturbation  generated by TenAd  adversarial attack for some frames}
	\label{fig:comp}
\end{figure*} 

\begin{table}[t]
	\centering
		\begin{tabular}{l|l|l}
			Attack & RE on UCF101 & RE on HMDB51 \\
			\hline 
			TLBBA & (1,1,1,1) & (1,1,1,1)  \\
			Heuristic\cite{heu} & (1,9,83,83) & (1,11,88,89) \\
			SVA\cite{sva} & (3,13,101,107) & (3,12,101,105)  \\
			VBAD\cite{vbad} &  (3,13,101,106) & (3,12,100,104) \\
		\end{tabular}
		\caption{Rank of error across three modes on C3D model and HMDB51.}
		\label{table2m}
	\end{table}
Additionally, as shown in Table~\ref{table2m}, our attack consistently generates low-rank adversarial examples compared to state-of-the-art methods. This outcome evidences our method's superiority, resulting in subtle adversarial perturbations that closely resemble the original videos, thus maintaining high perceptual similarity. 
\begin{figure}
	\centering
	\includegraphics[width=13cm]{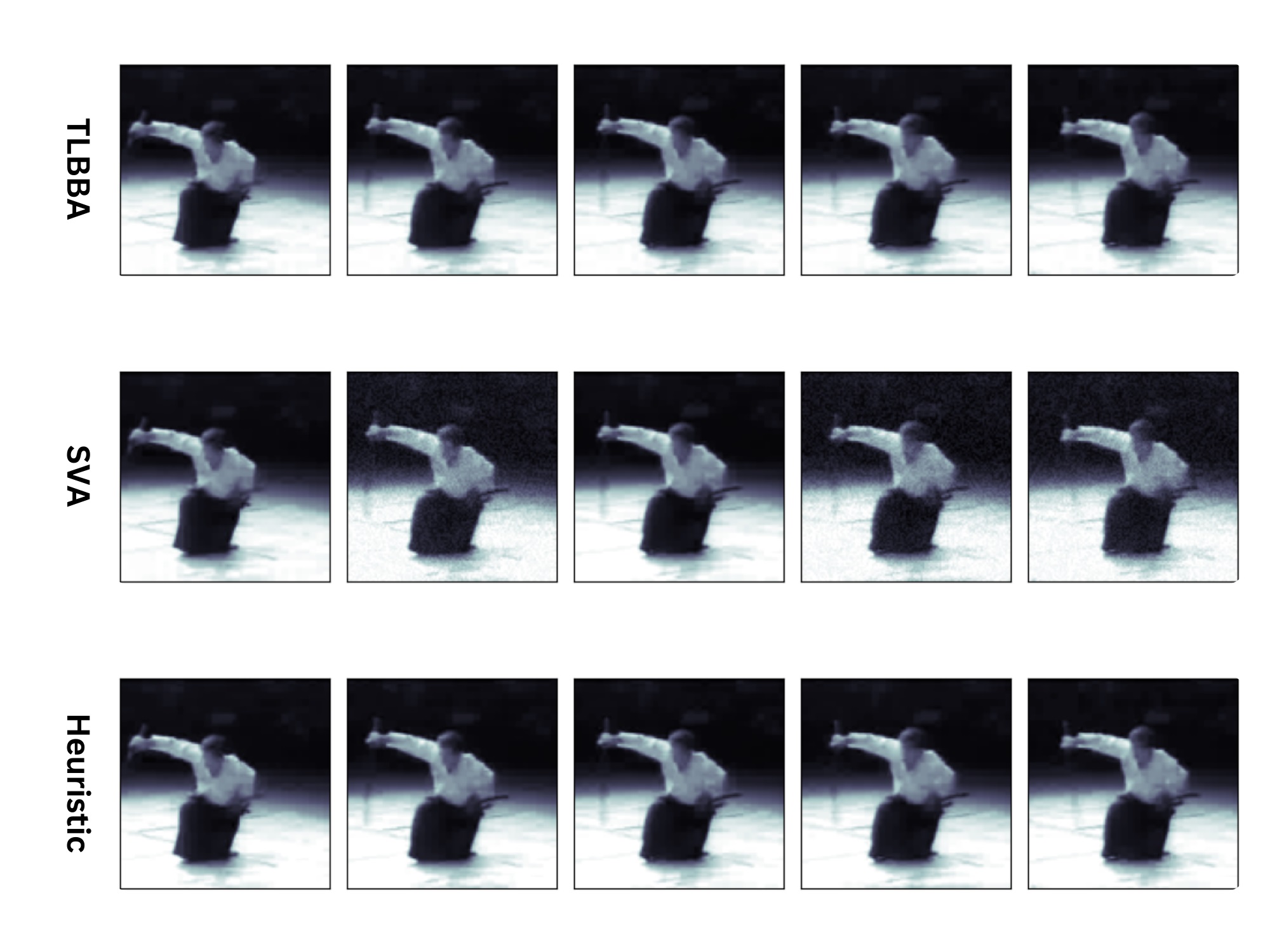}
	\caption{Adversarial frames generated by TenAd, Heuristic\cite{heu}, and SVA\cite{sva}}
	\label{fig:comp}
\end{figure}
Also, Figure \ref{fig:comp}, show some  examples of provided adversarial attacks by different methods.
 Our low-rank attack approach adeptly generates adversarial examples with remarkable similarity to clean videos by incorporating subtle low-rank perturbations. Notably, this impressive output is achieved without necessitating a substantial number of queries or prolonged time frames. In fact, our method yields a substantial reduction in the number of queries required—up to a 99\% reduction—rendering the attack virtually invisible to many security defenses. This result showcases the exceptional efficacy of our approach compared to state-of-the-art attacks.

Our evaluation encompasses various aspects, including the reduction in overall perturbation, significant decreases in query numbers and the time required for the attack, resulting in adversarial examples that are highly imperceptible to the human eye. We provide a detailed assessment of our method, showcasing its efficiency and effectiveness across multiple dimensions among state-of-the-art black-box attacks on video recognition models.
\section{Conclusion}
In this paper, we present a novel approach to black-box attacks on video models. We introduce TenAd, a tensor-based low-rank adversarial attack, where unlike the other methods uses the multidimensional structure of the video ata in attack. This technique produces low-rank perturbations, creating adversarial examples that are both query-efficient and time-efficient while maintaining high attack success rates. We validate the efficiency and effectiveness of our approach through experiments on the UCF101 and HMDB51 datasets.

 \bibliographystyle{elsarticle-num} 
 \bibliography{cas-refs}





\end{document}